# Emotion Analysis Platform on Chinese Microblog

Duyu Tang, Bing Qin, Ting Liu, Qiuhui Shi

Research Center for Social Computing and Information Retrieval Computer Science Department, Harbin Institute of Technology, China

{dytang, qinb, tliu, qhshi}@ir.hit.edu.cn

## 1 ABSTRACT

Weibo, as the largest social media service in China, has billions of messages generated every day. The huge number of messages contain rich sentimental information. In order to analyze the emotional changes in accordance with  time and space, this paper presents an Emotion Analysis Platform (EAP), which explores the emotional distribution of each province, so that can monitor the global pulse of each province in China. The massive data of Weibo and the real-time requirements make the building of EAP challenging. In order to solve the above problems, emoticons, emotion lexicon and emotion-shifting rules are adopted in EAP to analyze the emotion of each tweet. In order to verify the effectiveness of the platform, case study on the Sichuan earthquake is done, and the analysis result of the platform accords with the fact. In order to analyze from quantity, we manually annotate a test set and conduct experiment on it. The experimental results show that the macro-Precision of EAP reaches 80% and the EAP works effectively.

**Key Words**: Sentiment Analysis, Social Media, Emotion Analysis Platform, Weibo, Natural Language Processing

## 2 INTRODUCTION

The popularity of social media websites such as Twitter and Weibo has changed our life. Users can express their sentiments freely on these platforms. Weibo, as the most famous microblog service in China, has 503 million registered users until 2012, and approximately 100 million messages (known as tweet or weibo) are posted each day, according to the latest entry in Wikipedia (http://en.wikipedia.org/wiki/Sina_Weibo). Tweets may express emotions towards products, hot news or public events, so that mining sentiments from the big data of Weibo is practically useful. For example, consumers can seek advices from existing users before they purchase a product, the government can find the public opinions about their policies. Besides the sentimental information, Weibo also contains temporal and spatial information associated with each tweet. The temporal and spatial information makes it possible to analyze the global emotional changes according to time and space.

In this paper, in order to explore the emotion of the public and compare the sentiments of different area in real time, we propose an Emotion Analysis Platform (EAP)[1]. EAP can be applied in Weibo to monitor the emotional pulse of China, which aims to answer the following questions: (1) Can we rank the provinces by their happy degree via text analyzing on social media? (2)

[1] http://qx.8wss.com/

When a public event bursts, is it possible to analyze the emotion of tweets in real time to reflect people's emotional pulse in the real world?

The challenges of building EAP are mainly caused by two aspects, first, the huge data of Weibo and the real-time requirements, second, the texts of Weibo which have complex characteristics. Confronted with big data and real-time requirements, supervised methods are infeasible due to their dependence on annotated data. Compared with traditional review texts, texts on Weibo have some special characteristics which pose great challenges to sentiment analysis. The main challenges of Weibo sentiment analysis are as follows. First, the length of each tweet is up to 140 characters, which makes tweets extremely short and difficult to be analyzed. Second, the language style of tweets is informal. Slangs, abbreviations and elongated words, such as *coooollll,* are commonly used in tweets. It is difficult for the traditional parser to accurately identify the meaning of these tweets. Besides, abbreviations or acronyms will come up frequently on social media, which gives a big challenge to the robustness of the text analyzer. Finally, metadatas, such as hashtags and emoticons (also called smileys) are commonly used. In particularly, emoticons such as :) and :( are frequently used as sentimental indicators, which should be taken into consideration in the sentiment analyzer.

In order to understand the texts of Weibo without using parsers, the specific characteristic of tweets which can reflect sentiments are analyzed and utilized. The emoticons and hashtags of Weibo are fully utilized. Due to the massive data of Weibo and the real-time requirement, a rule-based method which combined emotion lexicon, emoticons and manually designed rules is proposed.

The EAP presents a general framework with three components, namely spider, emotion analyzer and user interface. The spider crawls tweets and sends them to the emotion analyzer. The emotion analyzer is the core component, which analyze a tweet as happy, sad, angry, surprise, fear or neutral. The sentiment analysis method is the rule-based method. The user interface makes the EAP easy to show the sentiment results. In order to verify the EAP from the senses, a case study of the earthquake in Sichuan province is done. The sentiment analysis results given by the EAP conforms to the real emotional change. In order to estimate the EAP's performance accurately, experiment is conducted on a data set which consists of 35,000 manually annotated tweets. The macro-Precision of EAP reaches 80%, and the experimental results verify the effectiveness of the EAP.

# 3 METHOD: EMOTION ANALYSIS PLATFORM

In this section, we first give the overview of EAP. Then, the detailed information of each component is followed.

The objective of EAP is to analysis people's emotion from social media in real time. To this end, three questions are needed to be answered: (1) How to collect data from social media in real time? (2) After the data is collected, how to detect the emotion of each case? (3) After the emotion of each instance is calculated, how to display the aggregated results of the province or the whole country?

In order to solve the above question, we split EAP into three components: Weibo Spider, Emotion Analyzer and User Interface (UI). The architecture of EAP is illustrated in Figure 1. The spider crawls tweets from Weibo online. Then, the tweets are fed into the Emotion Analyzer and each tweet will be classified as happy, sad, angry, surprise, fear or neutral. After accumulated by hour, the emotion distribution of each province will be grouped. Finally, the UI component shows the results on the website.

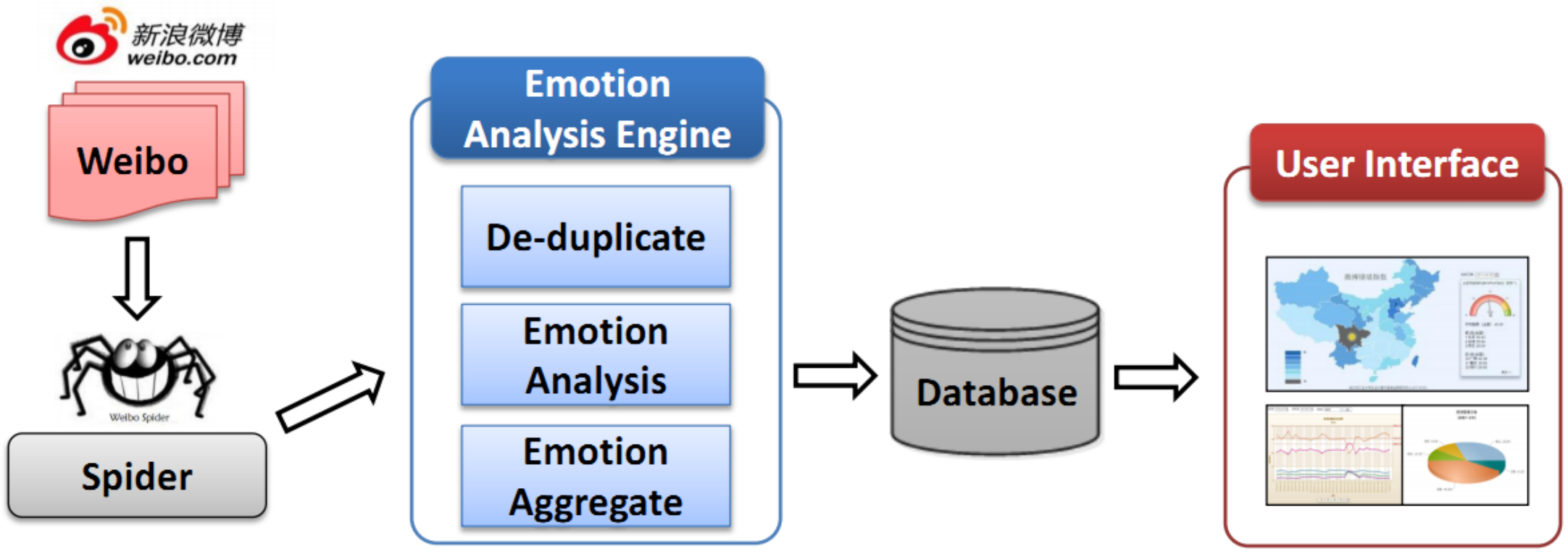

*Figure 1. Framework of Emotion Analysis Platform*

## 3.1 WEIBO CRAWLER

Data collection is the basic and critical component of EAP. Due to the objective of EAP is to analyze the texts on Weibo in real-time, the crawler must be designed robust to work online and diverse to sample data as similar as the real data distribution in the real world.

Through analysis, there are two alternative approaches to design Weibo Crawler. The first choice is to select active users from each province. Then, the tweets posted by them will be collected to calculate the emotion distribution of the corresponding province. However, there are three obvious shortcomings of this strategy. Firstly, the selected active users can not reflect the diversity of different classes of people in the real society, which makes the statistical results unreliable. Secondly, if the active user travels to another city, the count should be added into the emotional distribution of the target province. Thirdly, the active users may behave quietly in the social media over a period of time.

Based on the above consideration, we choose another strategy, which is to crawl tweets directly from the public-timeline of Weibo. The public-timeline shows the latest tweets from different provinces in real time. Then, each tweet crawled by our spider will be mapped into one of the 36 regions (34 provinces + 1 abroad + 1 other = 36). In Weibo, each tweet is optionally tagged with its posted location. If a tweet is attached with the location, we will map it to the tagged province. Otherwise, we will map the tweet into the province where the user registered. We utilize Weibo API (http://open.weibo.com/) to crawl tweets, within the limit of 200 tweets per second. After removing the duplicated tweets, we obtain about 3.2 billion tweets every day.

Through statistical analysis of the tweets on Weibo, the percentage of tweets from provinces in China is shown in Figure 2. From the figure, it can be seen that the distribution of tweet's amount in each province is consistent in two continuous months. In addition, four modernized metropolis (Guangdong, Beijing, Shanghai and Zhejiang) contribute most to the entire tweets.

## 3.2 EMOTION ANALYZER

As shown in Figure 1, tweets crawled by Weibo Spider are fed into the emotion analyzer. Then, emotion analyzer will tag each tweet as happy, sad, angry, fear, surprise or neutral. This subsection gives the detailed information about the algorithm used in emotion analyzer.

Emotion classification is essentially a text classification problem. Traditional sentiment classification mainly classifies documents as positive, negative and neutral. In this scenario, the emotion is fine-grained which contains happy, sad, angry, fear, surprise and neutral. Since it is a text classification problem, supervised [1] and unsupervised [2] methods are natural choices to be applied. Although supervised method is extensive studied for sentiment classification, it has two main weaknesses for practical usage. (1) Supervised methods usually require manually labeled corpus to train classifier, which is time-consuming and impractical in this scenario. (2) Supervised models are always domain-specific. However, the topics mentioned on tweets are diverse and variable over time. In this case, the classifier trained in the given domain or time period is not suitable for this “open” task.

Based on the above analysis, we choose to use unsupervised method to balance the effectiveness and the scalability of EAP as well as to maximize the characteristic of Weibo. Specifically, we utilize the combination of emoticons, emotion lexicons and emotion-shifting rules as the emotion analyzer. The rules are used together with emotion lexicons to count the frequency of each emotion. After added by another frequency distribution counted by emoticons, we choose the emotion with largest frequency as the final emotion. If all the emotional frequency are zero, the tweet is tagged as neutral. The emotion lexicon, emoticon and emotion-shifting rules used in the emotion analyzer will be described below.

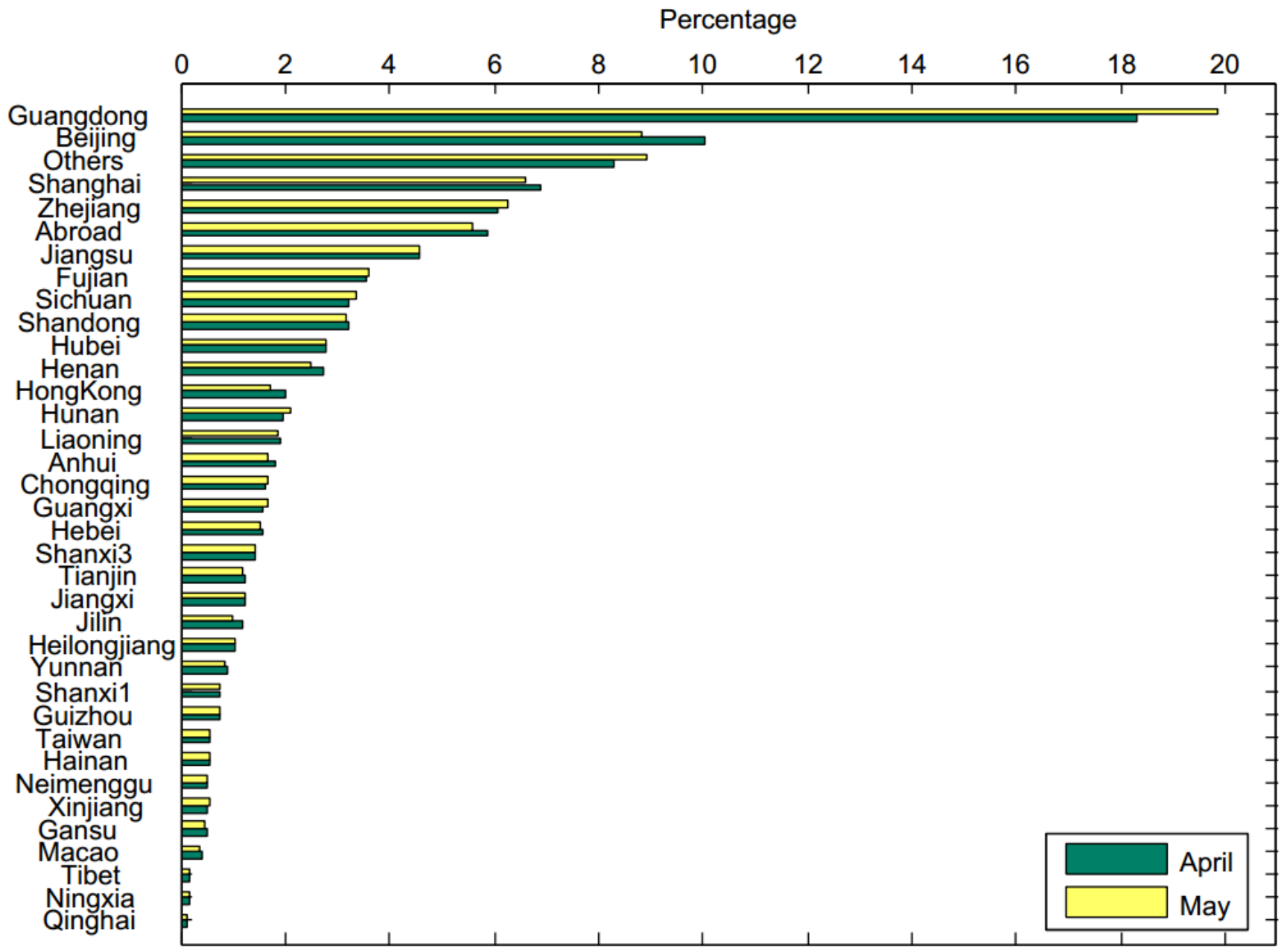

*Figure 2 Percentage of tweets from each province in China (April 2013 and May 2013).*

**Emotion Lexicon**

Peking Emotion Lexicon (EL is available at http://icl.pku.edu.cn/icl_res/) is used as the external lexicon resource. The detailed information about EL is given in Table 1.

*Table 1 Details of Emotion Lexicon*

| Emotion | Happy | Sad | Angry | Surprise | Fear |
|---|---|---|---|---|---|
| Lexicon-Size | 86 | 92 | 101 | 91 | 85 |
| Emoticon-Size | 8 | 8 | 3 | 4 | 2 |

**Emoticon**

In Weibo, there are more than one thousand official emoticons and only a small number of emoticons are indicative of emotions. Thus, we need to rank all the emoticons for each category to select the most representative and unambiguous ones. Inspired by the work of Li [3], the importance of each emoticon j in each emotion category i is calculated as shown in the following equation.

$$S_i(e_j) = \frac{\sum_k co_freq(e_j, sw_{ik})}{\sum_k \sum_I co_freq(e_j, sw_{Ik})} \times \log_{10}(freq(e_j))$$

The first multiplier corresponds to the quality factor and the second multiplier indicates the quantity factor. $freq(e_j)$ in the second multiplier stands for the frequency of the emoticon $e_j$ in the corpus. And $co_freq(e_j, sw_{ik})$ refers to the frequency that the emoticon $e_j$ and the k-th emotional word $sw_{ik}$ in the i-th emotion category co-occur within a message in the corpus. Finally, according to the ranking results, the top ranked emoticons are manually selected for each emotion, as shown in Table 2.

*Table 2 Emoticons for each emotion*

| Emotion type | Selected emoticons |
|---|---|
| Happy | 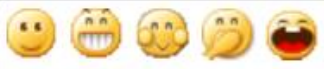 :) :-) :D |
| Sad | 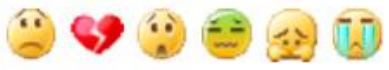 :( :-( |
| Angry | 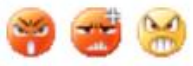 |
| Surprise | 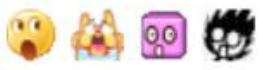 |
| Fear | 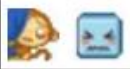 |

**Emotion-shifting Rules**

Sentiment shifters such as "no, not, never, none" can change the sentiment of text. For example, the emotion of sentence "I am not happy today." is sad because there is a "not" before "happy". In this work, we adopt three emotion-shifting rules, namely "Negation + happy -> sad; Negation + sad –> neutral; Negation + angry -> neutral".

We utilize the tweets from April and May in 2013 to conduct the preliminary study on emotion analyzer. Results are given in Figure 3. The two pies in the right hand give the ratio between emotional and non-emotional tweets. Emotional tweets are the tweets that recognized as happy, sad, angry, surprise or fear by emotion analyzer. The left pies show the distribution of the number of tweets in each emotion. It can be seen that, the distributions in two months are consistent in both settings. Moreover, happy is the dominant emotion in Weibo, almost half of the emotional tweets express happy feelings.

## 3.3 USER INTERFACE

In order to display the results analyzed by emotion analyzer, we design three web templates to visualize the results. They are (1) homepage, (2) global ranker and (3) province ranker. The homepage component mainly shows the happiness degree of each province on Chinese map, as illustrated in Figure 4. Each province will be filled in a kind of color. The happier province will be filled by darker blue. If the happy ratio of a province is under some threshold, its color will be filled as gray. The global ranker aims to rank the provinces by their happy ratio (Figure 5) and the province ranker gives the detailed emotional in the granularity of an hour (Figure 6).

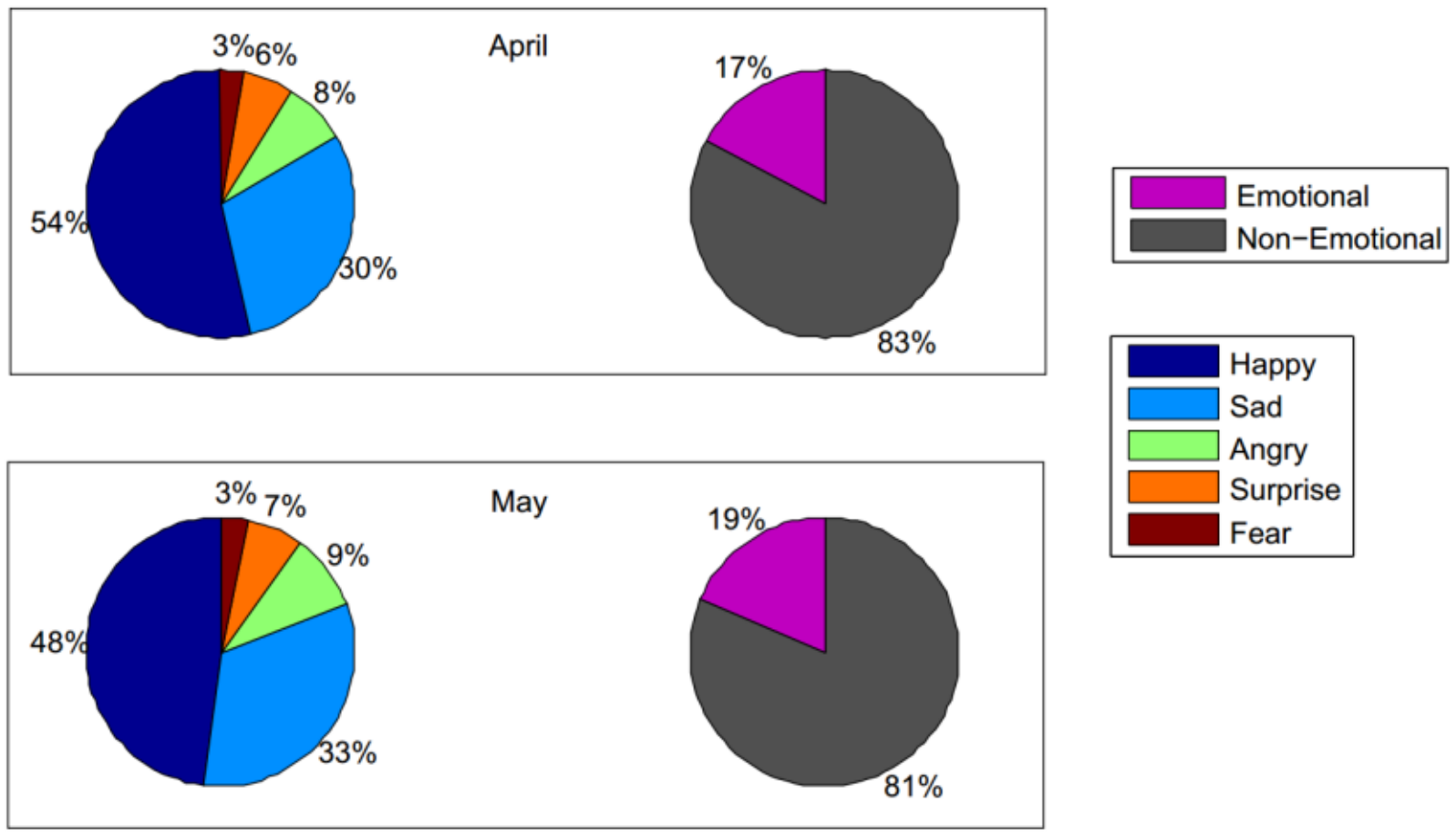

*Figure 3 Emotion Distribution of tweets on Weibo. (April 2013 and May 2013)*

# 4 RESULTS

In this section, we conduct qualitative and quantitative evaluation to check the performance of EAP.

## 4.1 CASE STUDY ON SICHUAN EARTHQUAKE

In this subsection, we give the qualitative evaluation by checking the results given by EAP when Sichuan earthquake occurred. Sichuan earthquake broke out at 8:00 am, 20$^{th}$ April in 2013 with ms 7.

The homepage of EAP at 20$^{th}$ April is given in Figure 4. At first glance, the color of Sichuan province is gray and attached with alarm, which indicates that the happy ratio in Sichuan province is extremely lower than the normal value. The dial on the right side shows that the average score of the whole country is 45.61, which is obviously lower than the average score in April. Besides, the province with lowest happy ratio is Sichuan and the penultimate province is Chongqing, adjacent to Sichuan province.

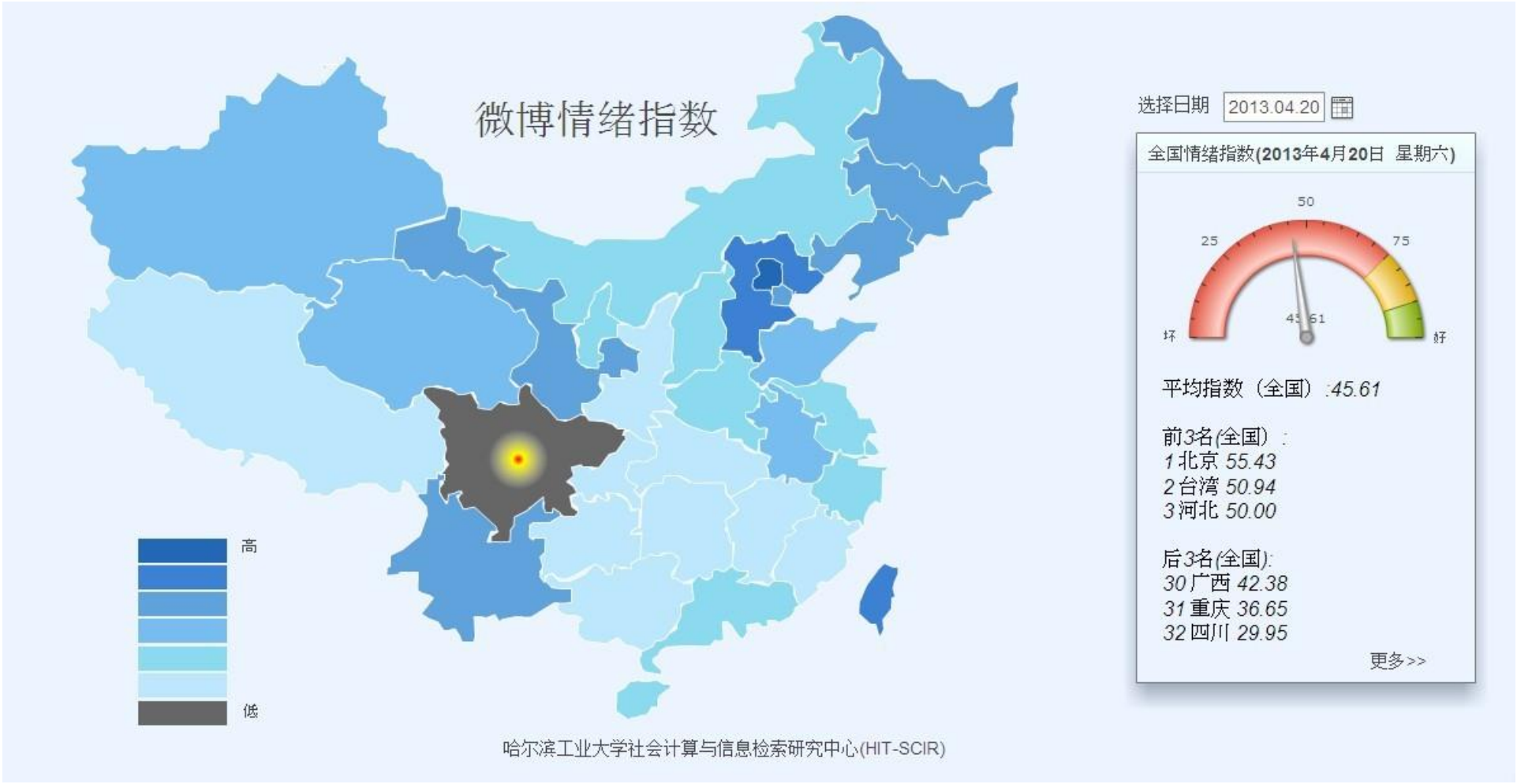

*Figure 4 The homepage of EAP at 20th April, 2013*

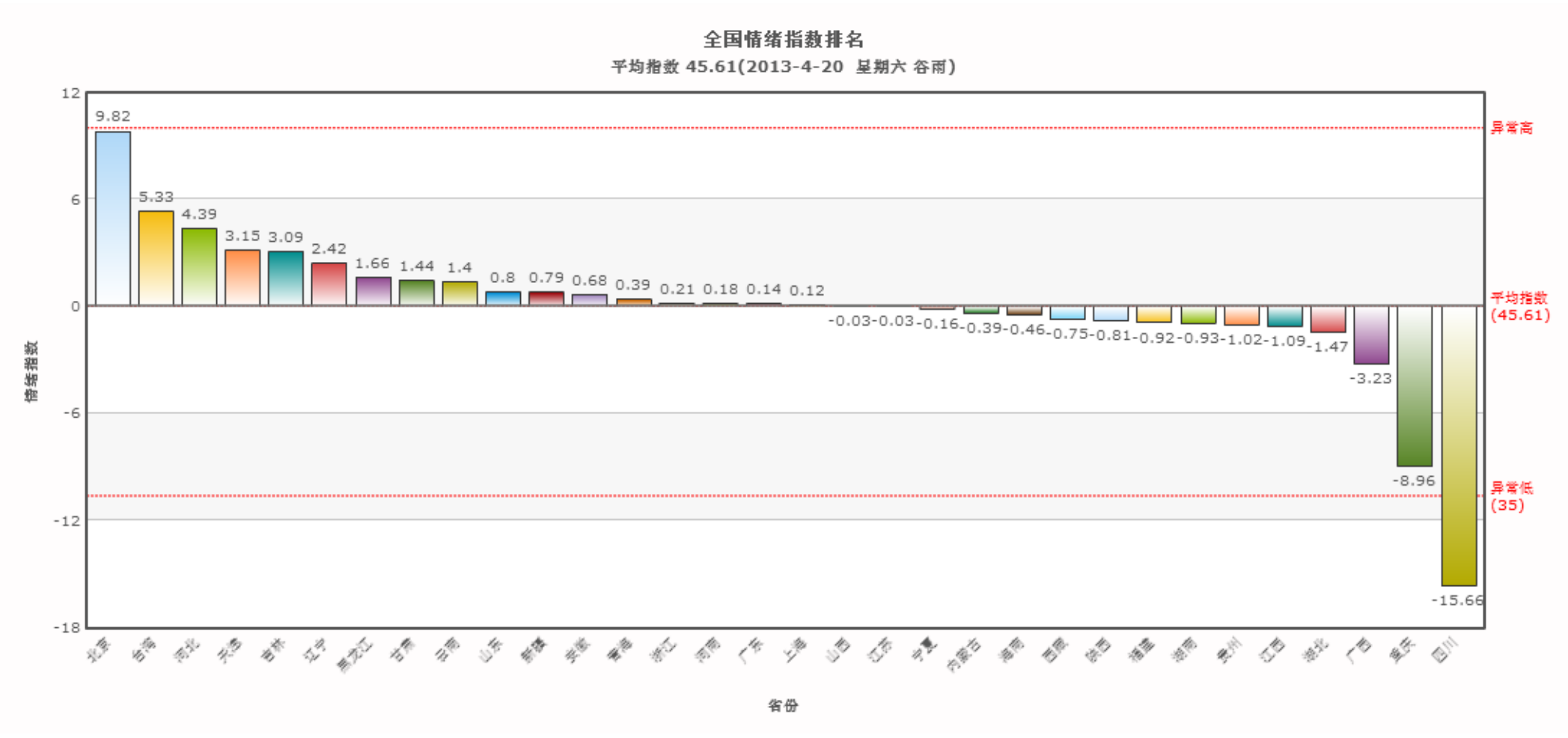

*Figure 5 The global ranker of EAP at 20th April, 2013*

The global ranker of EAP at 20th April is shown in Figure 5. The x axis stands for the provinces in China and the y axis stands for the happy ratio of each province minus the average happy ratio of all the provinces. These provinces are ranked by their happy ratios. It can be seen that the average score is 45.61 and the threshold for alarm is set to 35. Sichuan province is the one with lowest happy ratio, whose score is lower than the average score by 15.66. The score of its adjoining province, Chongqing, is lower than the average score by 8.96.

The province ranker gives the detailed emotion in the granularity of an hour. The result of Chongqing's province ranker at 20th April is shown in Figure 6. There are five lines with different colors, and each one corresponds to a kind of emotion. For example, the pink line means "happy" and the purple line means "fear". The x axis represents the 24 hours and the y axis stands for the emotional ratio of the each kind of emotion.

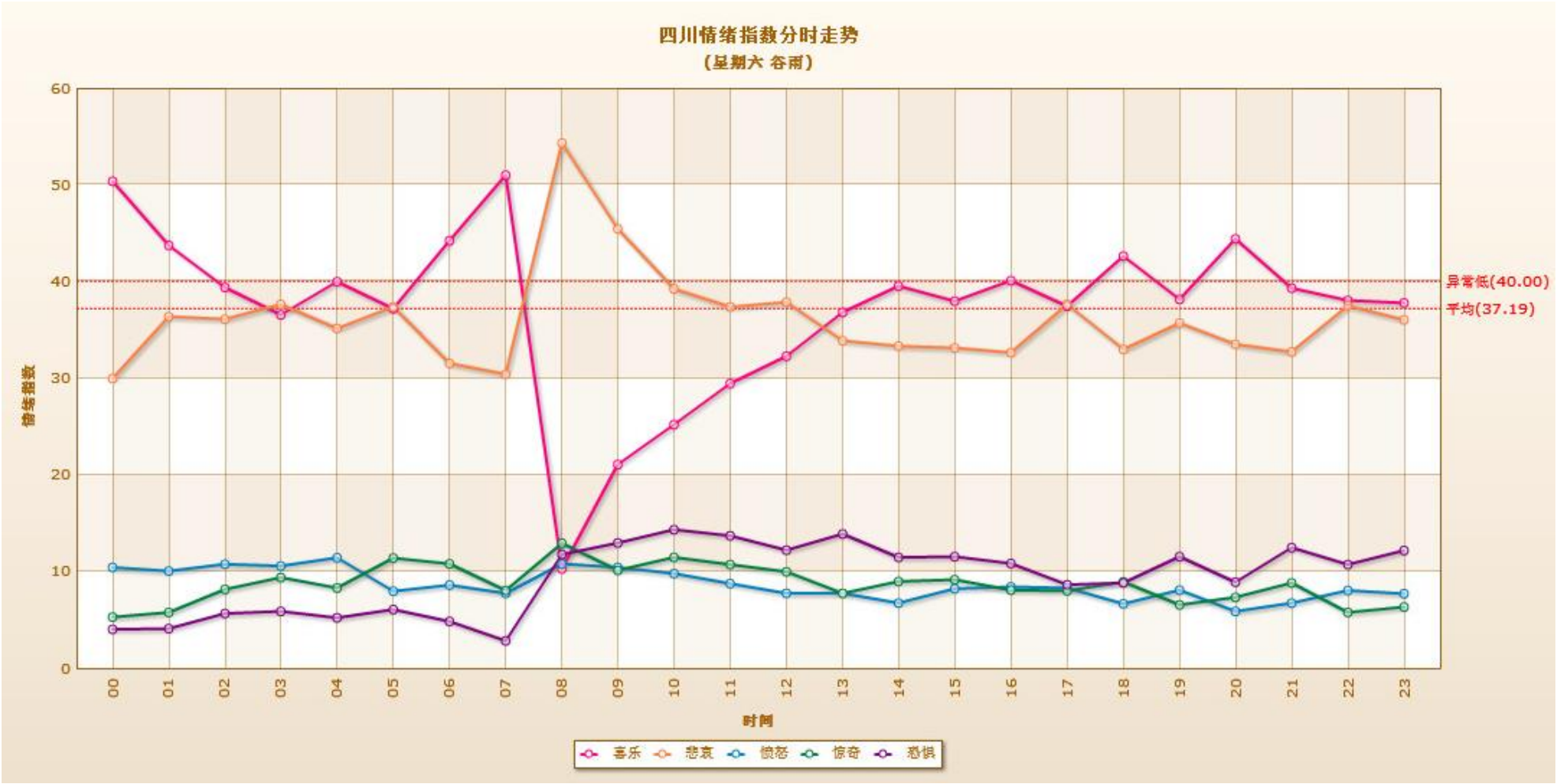

*Figure 6 The province ranker of EAP at 20th April, 2013*

It can be seen that the “happy” line has a sharp decrease at 8:00 am, just the hour that the earthquake strikes, and the ratios of “fear” and “sad” increase at the same time. After 8:00 am, “fear” remains while “happy” increases and “sad” decreases slightly. After analyzing the emotional tweets on 20th April, we find that the emotional recovery of “happy” after 8:00 am is due to the heartening words to Sichuan, such as “Cheer up! Sichuan!”, “Hope everyone well in Sichuan”.

## 4.2 Evaluation on Precision

In order to conduct the quantitative evaluation of EAP, we manually check the precision of tweets that tagged as emotional by EAP. We randomly select tweets from two continuous weeks. For each day and each emotion, 500 tweets are randomly selected to be checked that whether the tweet is tagged with the correct emotional label. Thus, the whole test set contains about 35,000 tweets (5 emotions * 7 days * 2 weeks * 500 tweets = 35,000). The precision of each kind of emotions is listed in Table 3. The macro-Precision of the five categories is 80%, which reflects the effectiveness of EAP.

*Table 3 Precision of EAP*

| Emotion | Happy | Sad | Angry | Surprise | Fear |
|---|---|---|---|---|---|
| Precision | 81.0 | 80.0 | 82.0 | 84.0 | 72.5 |

# 5 Related Work

With the popularity of blogs and social media, sentiment analysis (or opinion mining) has become a hot point in natural language processing research community [4-7]. This section will review prior studies from technical (subsection 6.1) and practical (subsection 6.2) perspectives, respectively.

## 5.1 EMOTION ANALYSIS AND TWITTER SENTIMENT ANALYSIS

The original attempt of sentiment analysis [1-2] aims to classify whether a whole document expresses a positive or negative sentiment. Pang [1] treats the sentiment classification of reviews as a special case of text categorization problem and first investigates machine learning methods. In their experiments, the best performance is achieved by SVMs with bag-of-words representation. Apart from supervised method, unsupervised methods [2, 8] have also been proved to be effective.

Apart from positive and negative evaluations, some researchers aim to identify the emotion of text, such as happy, sad, angry, etc. Mishne [9] uses emoticons labeled by the blogger to collect corpus in LiveJournal. And similar with [1], SVMs is utilized to train a emotion classifier with a variety of features over 100 emotions. Mishne [10] uses a similar method to identify words and phrases in order to estimate aggregate emotion levels across a large number of blog posts. Yang [11] combines SVMs and CRF for emotion classification at the document level. As social media become popular, Twitter sentiment analysis attracts much researchers' attention. Go [12] collects positive and negative data automatically with emoticons such as :-) and :-(. Kouloumpis [13] and Davidov [14] go further and use both hashtags and smileys to collect corpus. In addition, they use a KNN-like classifier for multiple emotion classification. Barbosa [15] leverages three sources with noisy labels as training data and use a SVM classifier with a set of features. From a different perspective, Liu [16] trains a language model based on the manually labeled data, and then use the noisy emoticon data for smoothing.

## 5.2 PLATFORM FOR SENTIMENT ANALYSIS

Traditionally, the majority of sentiment analysis applications focus on the area of reviews from consumer products and services. There are many websites that provide aspect-based opinion summary. For example, given the product reviews on one cell phone, the sentiment platform will give the detailed summary of reviews from the perspective of voice, screen, battery, weight, and so on [17]. Some applications can also provide the comparative results between two different products from the aspect-level. The representative platforms based on product reviews include Google Product Search (http://www.google.com.hk/shopping/product/), Bing Shopping (http://www.bing.com/shopping), and OpinionEQ (http://www.opinioneq.com/).

The explosion of social media services, such as Twitter and Weibo, presents a great chance to observe the public sentiment via analyzing the opinion-rich data. Many applications arise for the political, financial and commercial purpose. For political purpose, sentiment analysis on Twitter enables campaign managers to track how voters feel about the candidates (http://www.nytimes.com/interactive/us/politics/2010-twitter-candidates.html). For financial usage, the Stock Sonar (http://www.thestocksonar.com/) shows the daily positive and negative sentiment about each stock alongside the graph of the price of the stock. Similarly, the daily live stock market is predicted and tracked using Twitter sentiment analysis (http://liu.cs.uic.edu/TwitterWeb/DJIA_vs_opinion_with_TimeLine.jsp?type=mood). For commercial usage, Tweetfeel (http://www.tweetfeel.com) can perform real-time analysis of tweets to monitor the reputation of a brand. Moreover, there are applications for monitoring the pulse of global sentiment, such as Global Pulse (http://www.unglobalpulse.org/).

# 6 CONCLUSION AND FUTURE WORK

In this study, we present an Emotion Analysis Platform (EAP) on Chinese microblog. EAP can be applied in social media to capture the pulse of the global emotion from the dimensions of time and space. The case study on Sichuan earthquake shows that the global emotion distribution and the variability of fine-grained emotions by the hour given by EAP accord with the fact. Furthermore, the quantitative evaluation shows that the macro-Precision of EAP reaches 80%, which verifies the effectiveness of EAP.

As to future work, our plan is to design an event-dependent sentiment analysis platform on microblog based on EAP. This is especially meaningful to observe the public sentiment focusing on some breaking news.

# 7 ACKNOWLEDGEMENT

This work was supported by National Natural Science Foundation of China (NSFC) via grant 61133012, NSFC via grant 61073126 and NSFC via grant 61273321. We thank Jianfei Guo and Haochen Chen for their great help to build EAP. We thank Yaming Sun for refining the paper writing. We thank Ruiji Fu for helpful advice on UI design.